\documentclass[journal]{IEEEtran}

\usepackage{times}
\usepackage{epsfig}
\usepackage{graphicx}
\usepackage{amsmath}
\usepackage{amssymb}
\usepackage{enumitem}
\usepackage{array}
\usepackage{multirow}
\usepackage{subfigure}
\usepackage{amsthm,amsmath,amssymb}
\usepackage{mathrsfs}
\usepackage{booktabs}
\usepackage{cite}
\usepackage{color}

\ifCLASSINFOpdf
\else
\fi

\begin{document}

\title{EPG-MGCN: Ego-Planning Guided Multi-Graph Convolutional Network for Heterogeneous Agent Trajectory Prediction}

\author{\IEEEauthorblockN{Zihao Sheng, Zilin Huang, Sikai Chen}
\thanks{{\it Corresponding author: Sikai Chen.}}
\thanks{The authors are with the Department of Civil and Environmental, University of Wisconsin-Madison, Madison, WI 53706, USA (E-mails: \{zihao.sheng, zilin.huang, sikai.chen\}@wisc.edu).}}

\maketitle

\begin{abstract}
  To drive safely in complex traffic environments, autonomous vehicles need to make an accurate prediction of the future trajectories of nearby heterogeneous traffic agents (i.e., vehicles, pedestrians, bicyclists, etc). Due to the interactive nature, human drivers are accustomed to infer what the future situations will become if they are going to execute different maneuvers. To fully exploit the impacts of interactions, this paper proposes a ego-planning guided multi-graph convolutional network (EPG-MGCN) to predict the trajectories of heterogeneous agents using both historical trajectory information and ego vehicle's future planning information. The EPG-MGCN first models the social interactions by employing four graph topologies, i.e., distance graphs, visibility graphs, planning graphs and category graphs. Then, the planning information of the ego vehicle is encoded by both the planning graph and the subsequent planning-guided prediction module to reduce uncertainty in the trajectory prediction. Finally, a category-specific gated recurrent unit (CS-GRU) encoder-decoder is designed to generate future trajectories for each specific type of agents. Our network is evaluated on two real-world trajectory datasets: ApolloScape and NGSIM. The experimental results show that the proposed EPG-MGCN achieves state-of-the-art performance compared to existing methods.
\end{abstract}

\begin{IEEEkeywords}
  Heterogeneous trajectory prediction, multi-graph convolutional network, autonomous driving.
\end{IEEEkeywords}

\section{Introduction}
\IEEEPARstart{A}{utonomous} vehicles have the potential to enhance the safety, mobility, and comfort of the transportation systems, because they, unlike human drivers, can continually observe nearby traffic agents' behaviors and plan safe and efficient future motions without being distracted or tired \cite{hou2019interactive,muhammad2020deep,chen2021graph}. 
While great progress in artificial intelligence and robotics has been made in the past decades, many challenges still remain to achieve fully autonomous driving. One of the critical challenges is that autonomous vehicles need to accurately predict motion behaviors of nearby heterogeneous traffic agents, so as to ensure safety and plan future motions reasonably in complex traffic environments \cite{cui2021lookout,messaoud2021attention,sheng2022graph}.

The trajectory prediction for traffic agents is a challenging task, considering that so many latent variables affect the future trajectories. 
Firstly, the future motions of a traffic agent are highly dependent on the complex interactions with nearby agents. In the spatial dimension, agents at different locations have different impacts on other agents. In the temporal dimension, the same agent might have different impacts on others at different timestamps. 
Secondly, the actions to be performed by the ego vehicle will affect the future behaviors of surrounding traffic participants, resulting in different prediction results. Given multiple future paths of the ego vehicle, nearby traffic agents would have different reactions to them. 
Thirdly, when driving in urban traffic, different types of road users may coexist, such as cars, bicyclists, pedestrians. Due to the differences in their shapes and motion patterns, it is necessary to distinguish their category-specific behaviors and interactions.

In spite of these challenges, great progress in improving the accuracy of trajectory prediction has been made by recent studies that focus on deep learning methods. With the success of long short-term memory (LSTM) networks in capturing the temporal relations \cite{ma2015long,greff2016lstm}, pioneering works \cite{alahi2016social,deo2018convolutional,zhao2019multi} proposed LSTM-based networks to predict future vehicle trajectories. 
However, LSTM is inefficient and nonintuitive to model the social interactions (i.e., how individuals act and react to each other) as spatial relations in the interactions have irregular structures \cite{mohamed2020social}. 
In order to solve these issues, researchers have made efforts on modeling the interactions between traffic agents using graph convolutional networks \cite{li2019grip,mohamed2020social,yang2021novel}. 
The performance of these methods is quite satisfying, but they were designed for one particular type of traffic agents. As a result, they may not work well in urban traffic scenarios where multiple types of traffic agents share the road and complex interactions exist. Additionally, these methods predict the trajectory from the perspective of a static observer that is not involved in traffic. However, when an autonomous vehicle drives on the road while predicting others' trajectories, it is necessary to consider the effects of its future plans on the prediction results.

Therefore, to overcome the above limitations, this paper proposes a ego-planning guided multi-graph convolutional network (EPG-MGCN) to predict future trajectories for multiple heterogeneous traffic agents simultaneously. First, we propose to model social interactions using multiple graphs from four perspectives, i.e., distance, visibility, ego planning, and category, to generate distance graphs, visibility graphs, planning graphs, and category graphs, respectively. 
The distance graphs and visibility graphs consider the fact that a moving agent would pay more attention to other agents that are closer to it and in its vision field, respectively. The planning graphs describe the impacts of the ego vehicle's planning information on the future trajectories of surrounding agents. The category graphs utilize the similarity in motion patterns of the same type of agents to improve prediction results.
Besides, to distinguish the asymmetric effects between traffic agents, the visibility graph and planning graph are modeled as directed graphs. 
Then, the planning information of the ego vehicle is encoded and concatenated with features extracted by the multi-graph convolutional network. This further endows our network with the capability to learn how other agents will interact with the ego vehicle's planned trajectories. 
Finally, to deal with the heterogeneity of traffic agents, a category-specific gated recurrent unit (CS-GRU) encoder-decoder is designed, which generate future trajectories for each specific type of agents. The experimental results on complex traffic scenarios demonstrate the superior performance of our proposed model when compared with existing approaches. The main contributions of our proposed network are as follows:

(1) The planning-guided prediction module is proposed to encode the future planning information of the ego vehicle. The future plans inform the relationship between the agents to be predicted and the ego vehicle, and thus alleviate the uncertainty from the driving behaviors, leading to an improvement in prediction accuracy.

(2) By resorting to multiple graph topologies, our proposed network well describes the social interactions between heterogeneous traffic agents from four perspectives. This provides a basis for more accurate modeling, and therefore, improves the prediction accuracy. Also, we propose two directed graph topologies to model the asymmetric effects between each two agents.

(3) Considering that the characteristics of motion patterns are similar for traffic agents belonging to the same type, and they usually vary among different agent types, a category graph and category-specific GRU are proposed to handle the heterogeneity of traffic agents.

The rest of the paper is structured as follows. In Section II, we review the related works. The problem formulation is given in Section III. Section IV presents a detailed description of the proposed network. In Section V, experimental results and analysis are presented. Finally, conclusions and future works are elaborated in Section VI.

\section{Related Works}
In the past few years, the increasing interest in autonomous driving makes the research community pay much attention to trajectory prediction. In this section, we give a brief review of related works.

\subsection{Trajectory Prediction Methods} 
Traditional approaches on trajectory prediction use hand-crafted features or cost functions to model interactions and constrains between vehicles \cite{choi2012unified,bahram2016combined,deo2018would}.
These approaches have high computational efficiency, but the performance of these approaches will degrade when encountered with un-predefined traffic scenarios.
Other traditional methods, including polynomial fitting \cite{houenou2013vehicle}, Hidden Markov Models \cite{firl2012predictive,deo2018would} and Bayesian networks \cite{lefevre2011exploiting}, have been studied for this task, but they can only tackle traffic scenarios with simple interactions.

Nowadays, deep-learning-based-models for trajectory prediction \cite{deo2018convolutional,sadeghian2018car,deo2018multi} outperform the above mentioned traditional approaches as they can learn to extract features automatically from abundant realworld data.
Social-LSTM \cite{alahi2016social} is one of the pioneering works that apply deep learning to handle the task of human trajectory prediction. The authors proposed a social pooling mechanism to learn the interactions among pedestrians. 
Zyner {\it et al.} \cite{zyner2018recurrent} proposed an LSTM-based model to predict a potential direction that the driver would take at an intersection.
Similarly, Xin {\it et al.} \cite{xin2018intention} proposed a dual LSTM-based model to estimate driver intentions and predict future trajectories.
Later on, inspired by the interactive nature of road users, researchers began to consider the effects of nearby agents to design new deep-learning-based models.
For instance, Deo {\it et al.} \cite{deo2018convolutional} proposed a model named CS-LSTM, in which a convolutional social pooling was introduced to learn effects of nearby vehicles on the target vehicle.
Similarly, Zhao {\it et al.} \cite{zhao2019multi} proposed a multi-agent tensor fusion (MATF) network to use CNNs to extract scene features from the scene images and then put the encoded historical trajectories of all vehicles on the scene feature map to model the spatial interactions.
MATF also takes the advantage of generative adversarial networks (GANs) to learn the complex interactions among vehicles.
These approaches achieve a higher trajectory prediction accuracy than traditional methods. However, they were designed for one type of road users and may not work well for the others.

Recently, to improve the safety of autonomous driving in urban environments, researchers make effort on studying how to accurately predict future trajectories for heterogeneous traffic participants with a unified network. 
Ma {\it et al.} \cite{ma2019trafficpredict} proposed a LSTM-based model, which learns traffic agents' motions, interactions, and similarities within the same type to improve the prediction results.
Chandra {\it et al.} \cite{chandra2019traphic} proposed a LSTM-CNN hybrid network to model interactions, which implicitly accounts for the different shapes, motion pattens, and behaviors of traffic agents.
However, these works rely on LSTM or CNN to capture the interactions, which is inefficient.
Later on, Li {\it et al.} \cite{li2019grip} proposed a graph-based model GRIP++ to model the interactions between different traffic agents, improving the accuracy of the trajectory prediction for heterogeneous traffic agents. Mo {\it et al.} \cite{mo2022multi} proposed an edge-enhanced graph attention network to deal with the heterogeneity of the target agents and other road users.

Although the above deep-learning-based approaches have made great progress, they predicted trajectories from the perspective of a static road-side observer and thus ignored the influence of the ego-agent planning information on the future motions of surrounding traffic agents. The change of the ego vehicle's future driving path will affect the next behaviors of surrounding agents, and thus the prediction results should also change accordingly.
Inspired by this idea, Song {\it et al.} \cite{song2020pip} proposed a planning-informed trajectory prediction network (PiP) to incorporate planning information of the ego vehicle into prediction.
Guo {\it et al.} \cite{guo2022vehicle} improved PiP by applying spatial and temporal attention mechanisms to distinguish the effects of various encoded information on the prediction. However, these two approaches are based on LSTM, so they suffer from the same issues as mentioned above, i.e., being neither efficient nor intuitive in modeling the social interactions among vehicles.

\subsection{Multi-Graph Convolutional Networks} 
Abundant non-Euclidean data exists in traffic domain, so the convolution operations of traditional convolutional neural networks (CNNs) become less helpful when dealing with these data \cite{wu2020comprehensive,ye2020build}.
Graph convolutional networks (GCNs) extend the convolution operations defined on standard grids to graphs, so GCNs have a powerful ability to process data in non-Euclidean space \cite{kipf2016semi}. 
The works that apply graph to model traffic network have better performance in capturing the complex spatial dependencies than previous ones based on CNNs \cite{zhang2017deep,guo2019deep}.
Nowadays, GCNs have been widely used to handle the point/station prediction tasks in traffic domain and achieves the state-of-the-art performance, such as traffic flow prediction \cite{guo2019attention,zheng2020gman,zhao2019t}, travel demand prediction \cite{liu2019contextualized,tang2021multi} and human/vehicle trajectory prediction \cite{zheng2021unlimited,li2021hierarchical}.

Since one graph may not be able to describe pair-wise correlations between different agents comprehensively, multi-graph convolutional networks are further proposed to enhance the representation capacity of graphs \cite{chai2018bike,geng2019spatiotemporal,lv2020temporal,ke2021predicting,su2022trajectory}.
For example, Chai {\it et al.} \cite{chai2018bike} improved the precision of bike flow prediction by constructing a multi-graph network to capture three kinds of information, including distance, interaction, and correlation.
Geng {\it et al.} \cite{geng2019spatiotemporal} proposed a spatio-temporal multi-graph convolution network to forecast ride-hailing demand, which applied three graphs to model proximity, transportation connectivity, and functional similarity.
Su {\it et al.} \cite{su2022trajectory} proposed three directed graph topologies, i.e., view graph, direction graph, and rate graph, to forecast pedestrians' trajectories.
Given the effectiveness of multi-graph convolutional networks in related domains, we propose a novel multi-graph convolutional module, which contains distance graphs, visibility graphs, planning graphs, and category graphs, to enhance the performance of the proposed network.

\section{Problem Description of Trajectory Prediction}\label{sec:problem}
The scenario considered in this paper is that a controllable autonomous vehicle, denoted as the ego vehicle, can accurately perceive the motions of neighboring agents within a certain range, and then records their trajectories as discrete points.
The ego vehicle also has access to its planned future maneuvers and historical trajectories.
Given historical trajectories and the ego vehicle's planned future maneuver, we formulate the trajectory prediction as estimating future trajectories for a set of neighboring agents around the ego vehicle. These neighboring agents are called target agents as their trajectories need to be predicted.

To formulate this problem, we first introduce some notations. At a time $t$, the $i$th agent's historical trajectory over an observable time horizon $T_{obs}$ is denoted as,
\begin{equation}
  \mathbf{x}_i^t =[\mathbf{p}_i^{t-T_{obs}}, \cdots,\mathbf{p}_i^t],
\end{equation}
where $\mathbf{p}_i^t=(x_i^t, y_i^t)\in \mathbb{R}^2$ is a 2D position. The historical trajectories of the ego vehicle and all target agents are denoted as $\mathcal{X}= \{\mathbf{x}_i^t|i=0,1,\cdots,N\}$, where $N$ is the number of target agents and 0 denotes the index of the ego vehicle.
Similarly, the $i$th agent's future trajectory over a prediction time horizon $T_{pred}$ is represented as,
\begin{equation}
  \mathbf{y}_i^t = [\mathbf{p}_i^{t+1},\cdots,\mathbf{p}_i^{t+T_{pred}}].
\end{equation}
The set of all target agents' future trajectories is denoted as $\mathcal{Y}=\{\mathbf{y}_i^t|i=1,\cdots,N\}$. Note that the future trajectory of the ego vehicle is not included in $\mathcal{Y}$, as the ego vehicle is not required to be predicted.
Each target agent also has a category $c_i$ (e.g., vehicle, pedestrian, bicyclist). The set of all target agents' categories is denoted as $\mathcal{C}=\{c_i|i=1,\cdots,N\}$.
For the ego vehicle, we consider its planned future maneuver,
\begin{equation}
  \mathbf{y}_0^t = [\mathbf{p}_0^{t+1},\cdots,\mathbf{p}_0^{t+T_{pred}}].
\end{equation}

Thus, the trajectory prediction problem can be summarized as follows. Given the historical trajectories of the ego vehicle and all target agents, the category of each target agent, and the ego's planned future trajectory, the objective is to learn a posterior distribution $P(\mathcal{Y}|\mathcal{X}, \mathcal{C}, \mathbf{y}_0)$ of target agents' future trajectories.

\begin{figure*}
  \centerline{\includegraphics[width=17cm]{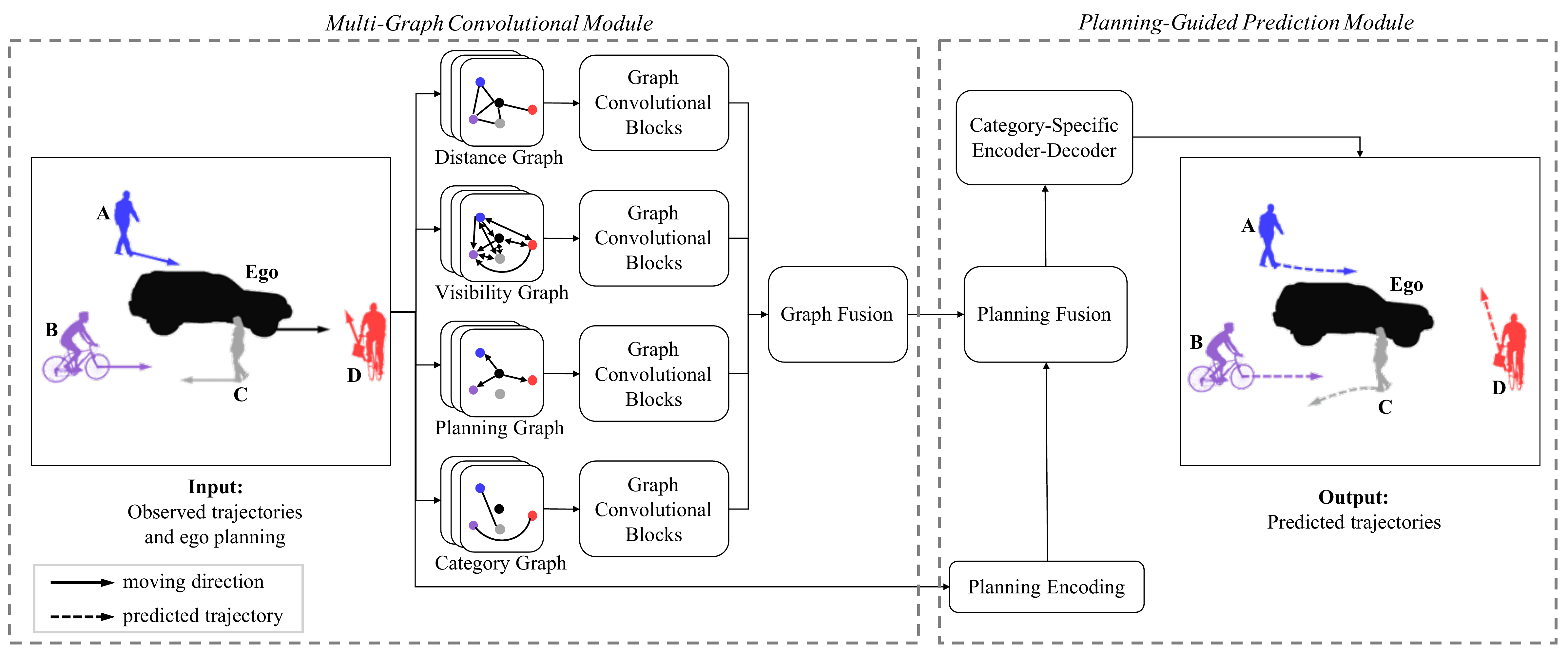}}
  \caption{Architecture of the proposed ego-planning guided multi-graph convolutional network (EPG-MGCN). The multi-graph convolutional module takes the observed trajectories as inputs and models the social interactions using four graphs, i.e., distance, vision field, ego planning, and agent category. The planning-guided prediction module first fuses the ego-agent plan with extracted graph convolutional features, and then sends it to the category-specific encoder-decoder to predict future trajectories (dashed lines) for each agent type.}
  \label{FG:frame}
\end{figure*}

\section{Ego-Planning Guided Multi-Graph Convolutional Network}
Previous graph-based trajectory prediction approaches usually only apply one undirected graph to model the social interactions. However, it is insufficient and unrealistic to describe the complexity and asymmetry in interactions. We propose to employ multi-graph and directed graph topologies to solve this issue.
Besides, traditional trajectory prediction methods usually predict traffic agents' future states from the perspective of a static road-site observer. 
 Nevertheless, autonomous vehicles are also dynamic traffic participants, and their planned next maneuvers will affect neighboring traffic agents' future motions. Therefore, we emphasize how the ego vehicle uses its planning information to make a more reasonable prediction.
The overall architecture of our proposed network is shown in Fig. \ref{FG:frame}. We introduce each component in the following subsections.

\subsection{Generation of Multiple Graphs}
Graph representation is an effective and intuitive technique for modeling the complex social behaviors among agents using data with irregular structures.
A graph can describe one type of relationship between each two agents.
However, agents in complex traffic usually exhibits multiple kinds of relationships, and thus a graph would fail to construct such complex impacts.
As a consequence, we propose a multi-graph convolutional module tailored for trajectory prediction.
We model the interactions between agents as multiple graphs by encoding the information of distances, visibility, planning, and category.

We first introduce the general definitions of graphs.
At a time step $t$, we define a graph as $G^t=(\mathcal{V}^t,\mathcal{E}^t)$, where $\mathcal{V}^t$ is the set of the ego vehicle and its nearby target agents.
Each $v_i^t\in \mathcal{V}^t$ represents an individual node, i.e., a traffic agent, and the attribute of $v_i^t$ is its coordinate $(x^t_i, y^t_i)$.
$\mathcal{E}^t=\{e_{ij}^t|i,j\in \mathcal{V}^t\}$ is the set of all weighted edges, where $e_{ij}^t\geq 0$. 
The larger the value of $e_{ij}^t$, the stronger the impact between agent $i$ and $j$.
A zero-value of $e_{ij}^t$ indicates there is no impacts between agent $i$ and $j$.
In the remaining, we omit the notation $t$ for convenience to introduce the definitions of each graph.

\subsubsection{Distance Graph}
Generally, a moving agent is more likely to be affected by the behaviors of a nearby agent rather than a distant one. 
In other word, an agent has different effects on other agents at different distances.
Following this idea, we construct a distance graph. Specifically, we use the reciprocal of the distance to mark the weight between two agents so that closer agents will be assigned with higher weights. 
The distance graph is denoted as $G^D=(\mathcal{V},\mathcal{E}^D)$. The element $e_{ij}^D \in \mathcal{E}^D$ is defined as,
\begin{equation}\label{Eq:dist}
  e_{ij}^D = \begin{cases}
    \frac{1}{\|\mathbf{p}_i-\mathbf{p}_j\|_2}, & \text{if } \|\mathbf{p}_i-\mathbf{p}_j\|_2 \leq d_D,
    \\0, & \text{otherwise},
  \end{cases} 
\end{equation}
where $\|\cdot\|_2$ denotes the $l_2$ norm, and $d_D$ is the distance threshold. Only when the distance is less than $d_D$ can two traffic agents be considered to have mutual impacts. As shown in Fig. \ref{FG:frame}, the bicyclist D is far away from the walker A and bicyclist B, so D has no edges connected to A and B in the distance graph.

\begin{figure}
  \centerline{\includegraphics[width=5.5cm]{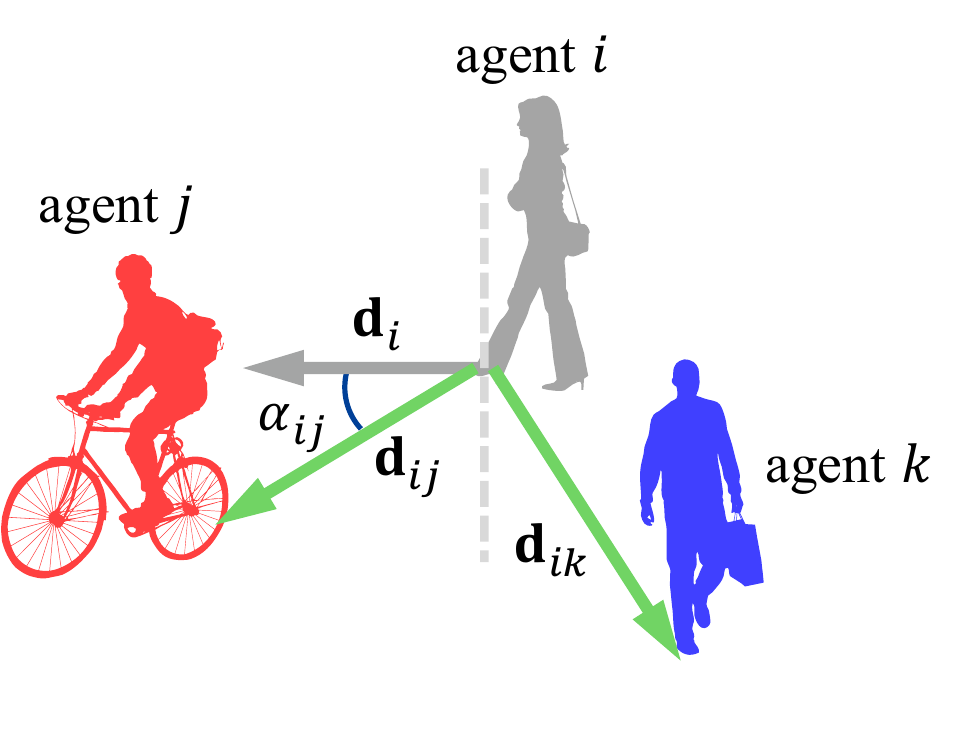}}
  \caption{Schematic plot of the impacts of vision field. The gray dashed line and solid line represent the boundary of views and moving direction of agent $i$, respectively. The green lines represent the relative position vectors from agent $i$ to other agents.}
  \label{FG:VG}
\end{figure}

\subsubsection{Visibility Graph}
The distance graph only exploits the distance information of agents. 
Intuitively, a moving agent would pay more attention to the agents in its vision field. 
Specifically, an agent's next action is more likely to be influenced by the agents in front as it can see their movements. On the contrary, it pays less attention to the behaviors of the agents behind it. Inspired by this prior, we propose the visibility graph to characterize such influence by resorting to the views of agents.
The view range for an agent is set as 180$^{\circ}$, and equally divided by its motion direction.
As shown in Fig. \ref{FG:frame}, the bicyclist D cannot see the bicyclist B, but B is in D's vision field. As a result, B has no impacts on D, but D can affect B.
The visibility graph is denoted as $G^V=(\mathcal{V},\mathcal{E}^V)$. The element $e_{ij}^V\in\mathcal{E}^V$ represents whether agent $j$ is in the view field of agent $i$.
We argue that when an agent appears in one's view field, the closer the distance is, and the closer the agent is to its visual center, the more attention it will pay to this agent. Thus, $e_{ij}^V$ is defined as,
\begin{equation}\label{Eq:view}
  e_{ij}^V = \begin{cases}
    \frac{\cos \alpha_{ij}}{\|\mathbf{p}_i-\mathbf{p}_j\|_2}, & \text{if } \mathbf{d}_i\cdot \mathbf{d}_{ij} > 0,
    \\0, & \text{otherwise},
  \end{cases} 
\end{equation}
where $\mathbf{d}_i=\mathbf{p}_i^{t+1}-\mathbf{p}_i^t$ represents the moving direction vector of agent $i$, $\mathbf{d}_{ij}=\mathbf{p}_j^t-\mathbf{p}_i^t$ denotes the relative position vector from agent $i$ to agent $j$, and $\alpha_{ij}$ is the angle between the moving direction $\mathbf{d}_i$ and relative position vector $\mathbf{d}_{ij}$, which is calculated as follows,
\begin{equation}
  \alpha_{ij} = \arccos (\frac{\mathbf{d}_i\cdot \mathbf{d}_{ij}}{\|\mathbf{d}_i\|_2\times \|\mathbf{d}_{ij}\|_2}).
\end{equation}
As illustrated in Fig. \ref{FG:VG}, we can see that $\mathbf{d}_i\cdot \mathbf{d}_{ij} > 0$ and $\mathbf{d}_i\cdot \mathbf{d}_{ik} < 0$, and thus the agent $i$ would pay attention to agent $j$ and ignore agent $k$.

\begin{figure}
  \centerline{\includegraphics[width=8cm]{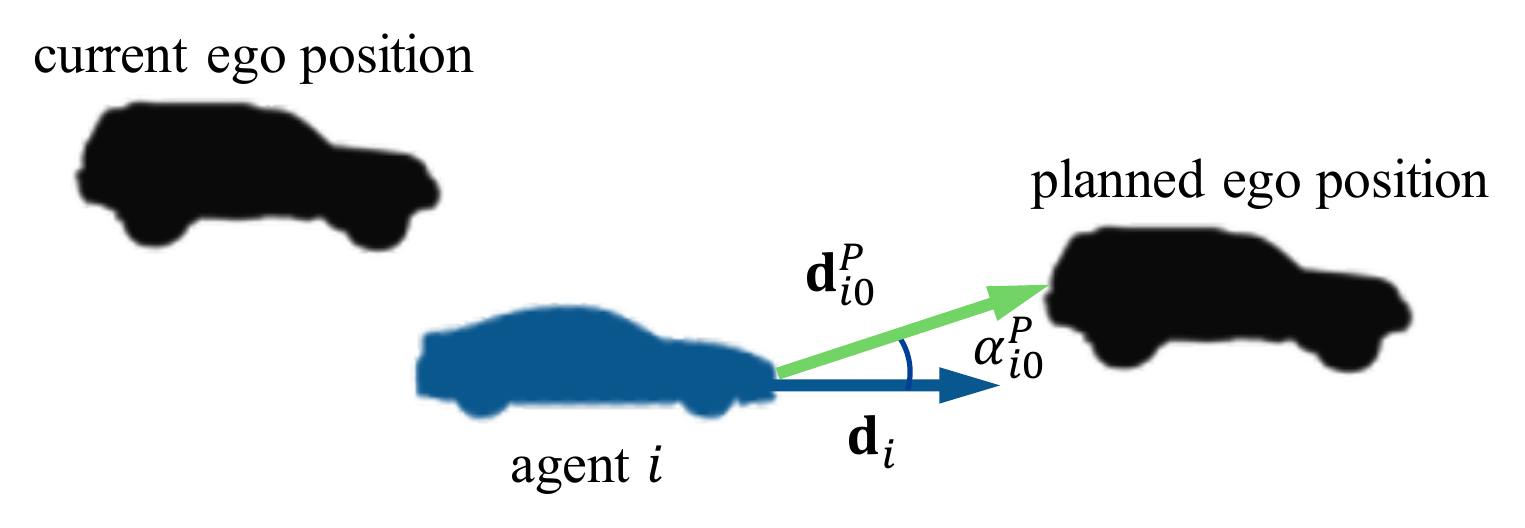}}
  \caption{Schematic plot of the impacts of ego-agent future plans. The darkblue line represents the moving direction of agent $i$. The green line is the relative position vector from agent $i$ to the planned ego future position.}
  \label{FG:PG}
\end{figure}

\subsubsection{Planning Graph}
The distance graph and visibility graph describe the distance and view field of traffic agents, respectively. However, in a crowded traffic scenario, the ego agent's planning information is also an important factor to influence others' future motions.
For example, at an unsignalized intersection, whether the autonomous vehicle stops will affect the decision of pedestrians to cross the crosswalk, leading to two completely different future situations.
In this paper, we propose the planning graph to exploit this prior. Specifically, one traffic agent would be impacted by the ego vehicle when the planned trajectory of the ego vehicle is within a range of $\pm \beta $ degrees regarding the speed direction of the traffic agent.
According to this rule, the planning graph is denoted as $G^P=(\mathcal{V},\mathcal{E}^P)$. The element $e_{ij}^P\in \mathcal{E}^P$ is defined as,
\begin{equation}\label{Eq:plan}
  e_{ij}^P = \begin{cases}
    1, & \text{if } j=0 \land \cos \alpha_{ij}^P \geq \cos \beta,
    \\0, & \text{otherwise},
  \end{cases} 
\end{equation}
where 
\begin{equation}
  \cos \alpha_{ij}^P = \frac{\mathbf{d}_i\cdot \mathbf{d}_{ij}^P}{\|\mathbf{d}_i\|_2\times \|\mathbf{d}_{ij}^P\|_2},
\end{equation}
$\mathbf{d}_{ij}^P=\mathbf{p}_j^{t+T_{pred}}-\mathbf{p}_i^t$ denotes the relative position vector from agent $i$ to agent $j$, and $\mathbf{p}_j^{t+T_{pred}}$ is the position of the ego vehicle at the last time-step in the prediction time horizon. 
As shown in Fig. \ref{FG:PG}, we can observe that the ego vehicle plans to overtake agent $i$, so the future motions of agent $i$ would be affected by the ego vehicle.

\subsubsection{Category Graph}
When making prediction for an agent in heterogeneous traffic, it is intuitive to refer to other agents that are similar to this one in terms of shapes and motion patterns. If two agents belong to the same category, we could think that they have a potential similarity in motion trends.
Therefore, we propose a category graph, denoted as $G^C=(\mathcal{V},\mathcal{E}^C)$, to exploit this prior. The element $e_{ij}^C\in \mathcal{E}^C$ is defined as,
\begin{equation}\label{Eq:category}
  e_{ij}^C = \begin{cases}
    1, & \text{if } c_i=c_j,
    \\0, & \text{otherwise},
  \end{cases} 
\end{equation}
where $c_i$ is the category of agent $i$, e.g., car, bicycle, pedestrian.

\subsection{Multi-Graph Convolutional Module}\label{sec:MGConv}
Based on the above four graph topologies, we introduce the framework of the multi-graph convolutional module in this subsection. 
As shown in Fig. \ref{FG:frame}, the multi-graph convolutional module consists of four parallel branches, each of which is designed to extracted spatial-temporal features from one of the above four graph topologies.
Each branch includes several graph convolutional blocks, and they perform graph convolutions in the spatial dimensions and 2D convolution operations in the temporal dimensions.
Assume that there are $N$ agents in a scene, we can construct four adjacent matrices $E\in\mathbb{R}^{N\times N}$ based on Eqs. (\ref{Eq:dist}), (\ref{Eq:view}), (\ref{Eq:plan}) and (\ref{Eq:category}), respectively.
The graph convolution operation is defined as:
\begin{equation}
\hat{Z}  = f(norm(\hat{E})Z{W}),
\end{equation}
where $Z\in \mathbb{R}^{N\times C \times T_{obs}}$ denotes the feature matrix of $N$ agents over $T_{obs}$ time steps, $f(\cdot)$ is an activation function, $\hat{E}=E+I$, $I\in \mathbb{R}^{N\times N}$ is an identity matrix, and ${W}$ is the learnable parameters matrix. $norm(\cdot)$ is to normalize the adjacent matrix and speed up the learning process of graph convolution \cite{kipf2016semi}, which is calculated as follows,
\begin{equation}
  \hat{e}_{ij} = \frac{e_{ij}}{\sum_{i=1}^N e_{ij}}.
\end{equation}
Note that the main difference of graph convolution operations in each branch lies in $E$.
The hop of graphs is set to be 2 to help capture information from distant neighborhoods.
After the graph convolution operations capture neighboring information in the spatial dimensions, we apply a 2D convolution layer, as reported in \cite{yan2018spatial}, to extract the temporal features by merging the information at the neighboring time slice. To apply convolution operations in the temporal dimensions, the kernel size of the convolutional layer is set to be $3\times 1$.

Each branch extracts different features $F_{dist}$, $F_{vis}$, $F_{plan}$, and $F_{cat}$, which contain  different priors of social interactions. 
Next, we fuse them into a unified feature. The fusion process is formalized as follows,
\begin{equation}
  F_{graphs} = Fuse(F_{dist}, F_{vis}, F_{plan}, F_{cat}),
\end{equation}
where $F_{graphs}\in \mathbb{R}^{N \times C \times T_{obs}}$ is the fused multi-graph feature. We first stack $F_{dist}$, $F_{vis}$, $F_{plan}$, and $F_{cat}$ to a tensor with a size of $4 \times N \times C \times T_{obs}$. Then, a convolution layer with $1\times 1$ kernel size is followed, and the $Relu$ activation function is used to obtain the fused graph feature $F_{graphs}$.

\subsection{Planning-Guided Prediction Module}
In \cite{li2019grip,sheng2022graph}, the target agents' future trajectories are directly decoded from the encoding features that only aggregates history information. 
In this way, each trajectory is generated from the corresponding target encoding without considering the ego vehicle's planning. However, the future states of each target agent are highly related with the planning of the ego vehicle. Therefore, in the planning-guided prediction module, we further consider the impacts of the ego vehicle's planning information on the trajectory prediction results.

Firstly, the ego agent's planned trajectory $\mathbf{y}_0\in \mathbb{R}^{2\times T_{pred}}$ is fed into a 1D convolution layer to obtain a temporal embedding with a size of $C\times T_{pred}$. 
Subsequently, this embedded trajectory information is further encoded by a GRU network to extract the temporal motion features, i.e., the hidden state with a size of $C\times 1$.
We then copy this planning encoding to a size of $N\times C\times T_{obs}$, and stack it with $F_{graphs}$ to a tensor of $2\times N\times C\times T_{obs}$. Similar to the graph fusion in Section \ref{sec:MGConv}, we use a convolution layer with $1\times 1$ kernels and $Relu$ activation function to fuse them. 
By fusing the information from the future plan of ego vehicle and the historical observation of neighboring agents, the resulting encoding $F_{fusion}\in \mathbb{R}^{N\times C \times T_{obs}}$ contains the spatial-temporal information from both the observation and prediction time domain.

After that, we design a category-specific GRU encoder-decoder network (CS-GRU) to deal with the heterogeneity in different kinds of traffic agents. 
The number of CS-GRU is equal to the type of traffic agents.
In this way, each CS-GRU takes $F_{fusion}$ as input to predict the future trajectories for only one type of traffic agents.
We take the operations on one agent as an example to elucidate the process of generating future trajectories from $F_{fusion}$.
The feature of this agent is extracted from $F_{fusion}$ and denoted as $F_{in}\in\mathbb{R}^{C\times T_{obs}}$.
The encoder GRU takes $F_{in}$ as input and generates output features and hidden state. 
The hidden state and the current position of this agent are fed into the decoder GRU as initial inputs to calculate the output features and hidden states.
Then, they are fed into the next GRU cell. We repeat this process until the trajectories in the whole prediction time horizon are generated.

\section{Experimental Evaluation}
In this section, we train and evaluate our proposed prediction model on two public trajectory datasets: ApolloScape and NGSIM. We implement the proposed model using the PyTorch framework \cite{paszke2017automatic} with an NVIDIA RTX 3080 GPU. 
Firstly, we compare the performance of our method against the state-of-the-art models quantitatively using the metrics of average displacement error (ADE) and final displacement error (FDE). 
Then, we conduct several ablation experiments to demonstrate the effectiveness of the proposed multi-graph convolutional module and planning-guided prediction module.
Finally, as our method could infer diffierent future situations by performing diffierent future ego plans under the same historical situation, we evaluate the rationality and diversity of the prediction results.

\subsection{Heterogeneous Agent Trajectory Prediction}
We first train our model on an urban traffic dataset to demonstrate its ability to handle heterogeneity of traffic agents.
\subsubsection{Dataset and Experimental Settings} 
The ApolloScape dataset \cite{ma2019trafficpredict} is a large-scale trajectory dataset in urban streets, which contains a total of 53 minutes of training set and 50 minutes of test set. All trajectories are annotated with a sampling frequency 2 Hz. Each scene contains up to 114 traffic agents belonging to 5 categories, i.e., small vehicles, big vehicles, pedestrian, motorcyclist/bicyclist, and others. For convenience, we treat small and big vehicles as one type.
Following existing works \cite{zhu2019starnet,carrasco2021scout,fang2020tpnet}, we use the first 3 seconds (6 frames) as the observation time horizon, and the next 3 seconds as the prediction time horizon.

When constructing distance graphs, we set the distance threshold $d_D$ to be 10 meters, i.e., any pair of traffic agents within 10 meters are connected using a weighted edge, as defined in Eq. (\ref{Eq:dist}). 
For the planning graph, we consider the ego vehicle's planned future motions would impact a traffic agent if the planned trajectory appears in the range of $\pm20^{\circ}$ regarding the speed direction of the agent, i.e., $\beta=20^{\circ}$. 
The batch size is set to be 128, and we train the model using Adam optimizer with an initial learning rate of 0.001. The learning rate is multiplied by 0.1 every 200 epochs until the convergence of the loss.

\subsubsection{Baselines} We evaluate the performance of our proposed method by comparing it with the following well-known baselines:

\begin{itemize}
  \item \textbf{TrafficPredict} \cite{ma2019trafficpredict}: This baseline is an LSTM-based network and proposes an instance layer and a category layer to learn interactions and heterogeneity. It is the baseline of the ApolloScape Trajectory Dataset.
  \item \textbf{Social GAN} \cite{gupta2018social}: This baseline uses GANs to learn the complex interactions among vehicles.
  \item \textbf{Social LSTM} \cite{alahi2016social}: This baseline is based on LSTMs and proposes a social pooling mechanism to learn the interactions among pedestrians.
  \item \textbf{StarNet} \cite{zhu2019starnet}: This baseline has a star topology, in which a unique hub network generates a description for interactions and multiple host networks predict future trajectories for each agent. 
  \item \textbf{GRIP++} \cite{li2019grip}: This baseline constructs a undirected graph to model the interactions among agents and uses LSTMs to predict future trajectories.
  \item \textbf{SCOUT} \cite{carrasco2021scout}: This baseline uses a undirected graph and three attention mechanisms to learn interactions, and proposes a noval loss to predict socially-consistent trajectories.
  \item Trajectory proposal network (denoted as \textbf{TPNet}) \cite{fang2020tpnet}: This baseline first predicts the ending points of target agents to produce multiple candidate trajectories, and then classify and refine these candidates to generate final predictions.
  \item Unlimited neighborhood interaction network (denoted as \textbf{UNIN}) \cite{zheng2021unlimited}: This baseline uses hierarchical graph attention to model category-to-category and agent-to-agent interactions.
\end{itemize}

\begin{table*}
  \caption{Comparison with baselines on the ApolloScape dataset. The best results are marked in bold. Data are in meters.}
  \label{TB:comp0}
  \centering
  \begin{center}
  \begin{tabular}{p{70pt}<{\centering}|p{32pt}<{\centering}|p{32pt}<{\centering}p{32pt}<{\centering}p{32pt}<{\centering}|p{32pt}<{\centering}|p{32pt}<{\centering}p{32pt}<{\centering}p{32pt}<{\centering}}
  \toprule
  Method & WSADE & ADEv & ADEp & ADEb & WSFDE & FDEv & FDEp & FDEb \\[1pt]
  \midrule 
  TrafficPredict & 8.59 & 7.94 & 7.18 & 12.88 & 24.23 & 12.77 & 11.12 & 22.79 \\[1pt]
  Social GAN & 1.96 & 3.15 & 1.33 & 1.86 & 3.59 & 5.66 & 2.45 & 4.72 \\[1pt]
  Social LSTM & 1.89 & 2.95 & 1.29 & 2.53 & 3.40 & 5.28 & 2.32 & 4.54 \\[1pt]
  StarNet & 1.34 & 2.39 & 0.78 & 1.86 & 2.49 & 4.28 & 1.51 & 3.46 \\[1pt]
  GRIP++ & 1.25 & 2.24 & 0.71 & 1.80 & 2.36 & 4.07 & 1.37 & 3.41 \\[1pt]
  SCOUT & 1.26 & 2.21 & 0.73 & 1.82 & 2.35 & 3.93 & 1.41 & 3.37 \\[1pt]
  TPNet & 1.28 & 2.21 & 0.74 & 1.85 & 1.91 & 3.86 & 1.41 & 3.40 \\[1pt]
  UNIN & 1.09 & - & - & - & \textbf{1.55} & - & - & - \\[1pt]
  \midrule
  EPG-MGCN (ours) & \textbf{0.96} & \textbf{1.58} & \textbf{0.62} & \textbf{1.29} & 1.58 & \textbf{2.65} & \textbf{1.01} & \textbf{2.09} \\[1pt]
  \bottomrule
  \end{tabular}
  \end{center}
\end{table*}

\subsubsection{Comparison of Prediction Results}
In this section, we present the comparison results between our proposed EPG-MGCN and the above baseline models.
In addition to ADE and FDE, the ApolloScape dataset also uses the weighted sum of ADE (WSADE) and weighted sum of FDE (WSFDE) as metrics to balance the different scales of different agent types.
The weights assigned to each type (i.e., vehicles, pedestrians and bicyclists) are 0.20, 0.58 and 0.22, respectively.
Table \ref{TB:comp0} lists the prediction accuracy of each method, where the values marked in bold represent the best results.
In the Table, the subscripts ‘v’, ‘p’, and ‘b’ denote agent types of vehicle, pedestrian, and bicycle, respectively.

According to the prediction results in Table \ref{TB:comp0}, we can see that our proposed EPG-MGCN achieves the best prediction results on all agent types. Moreover, the proposed method achieves the best or comparable performance in terms of the WSADE and WSFDE results. The excellent prediction results demonstrate the powerful ability of the EPG-MGCN in capturing the complex interactions among heterogeneous traffic agents and inferring their future trajectories.

When compared with GRIP++ and SCOUT based on single undirected graph topology, our EPG-MGCN gains over 23\% and 33\% performance improvement in terms of WSADE and WSFDE, respectively. This verifies that the proposed multi-graph topologies describing the impacts from multiple perspectives are indeed helpful for modeling the complex social interactions between traffic agents.

In addition, as Social GAN, Social LSTM, and StarNet are designed for the prediction task of homogenous traffic agents, we note that they have worse performance than those designed for heterogeneous agents (i.e., GRIP++, SCOUT, TPNet, and ours). Among all methods considering  heterogeneity, our EPG-MGCN is the only one that uses the planning information of the ego vehicle, and outperforms others that do not. Therefore, it is crucial to consider the ego-agent plans when predicting future trajectories of nearby agents.

\begin{figure*}
  \centerline{\includegraphics[scale=0.22]{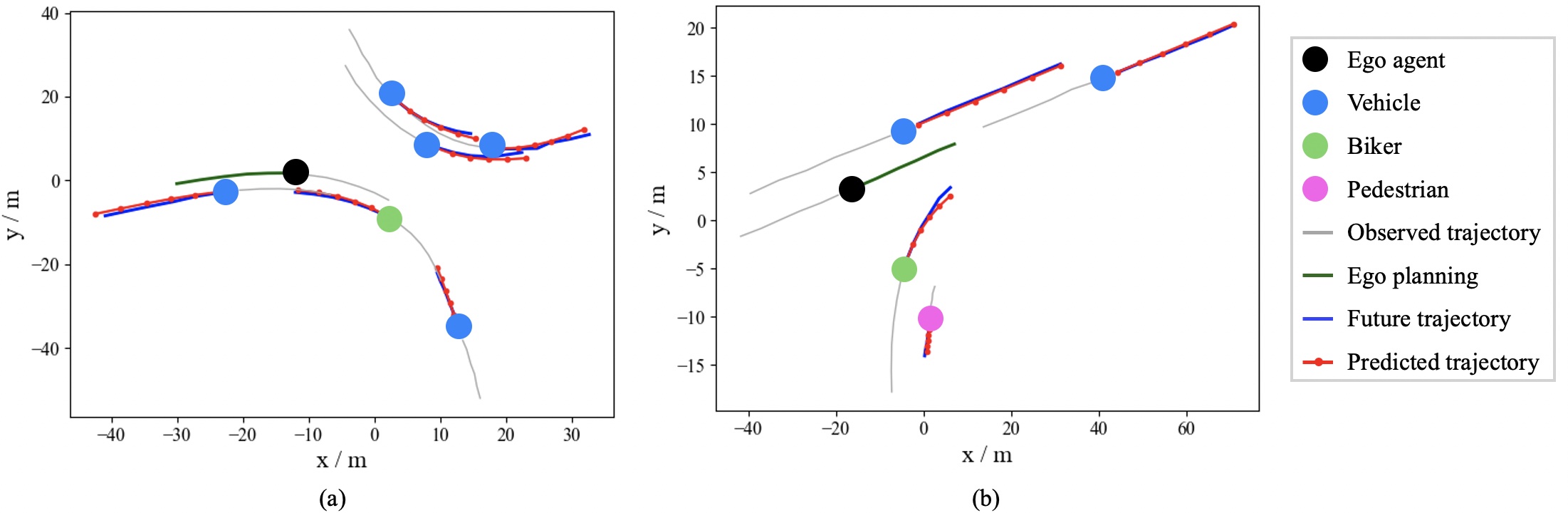}}
  \caption{Visualization of predicted trajectories on the ApolloScape dataset.}
  \label{FG:visual_apollo}
\end{figure*}

Next, we qualitatively analyze the prediction results by visualizing several representative predicted trajectories generated by our EPG-MGCN in Fig. \ref{FG:visual_apollo}. All the results are selected from the ApolloScape dataset. 
After collecting 3 seconds of historical trajectories and the future ego-agent plans, our network predicts their trajectories over the next 3 seconds. 
As shown in Fig. \ref{FG:visual_apollo}, we can see that the predicted trajectories can capture the evolving pattern of ground truth trajectories well on the whole prediction horizon.

Fig. \ref{FG:visual_apollo}(a) shows that there are two groups of traffic flow with opposite directions. Our proposed EPG-MGCN captures this information from multi-graph topologies, and thus accurately predicts the moving directions and speeds of surrounding heterogeneous agents.
As shown in Fig. \ref{FG:visual_apollo}(b), even if three kinds of traffic agents coexist on the road, our model still accurately predicts their future trajectories. It is worth noting that our network successfully predicts that the biker wants to turn to the right to avoid potential collision with cars.

\subsubsection{Ablation Study}
To understand why our EPG-MGCN has better prediction accuracy than others, we report the changes in WSADE while gradually adding the proposed components in Table \ref{TB:ablation0}, where $\checkmark$ indicates the corresponding component is used, and x indicates no such component in the model.
The adjustments we made are listed as follows:
\begin{itemize}
  \item Multiple graph topologies: We add the proposed four graph topologies, i.e., $G^D$, $G^V$, $G^P$, and $G^C$, to the proposed model one by one. 
  \item Planning-guided prediction (denoted as PGP): We verify the role of adding the ego-planning information to the planning-guided prediction module.
  \item Category-specific GRU (denoted as CS-GRU): Finally, we explore whether it is better to use category-specific GRUs or one generic GRU to handle the heterogeneity of traffic agents.
\end{itemize}

As shown in Table \ref{TB:ablation0}, we observe that from A1 to A6, the prediction accuracy is gradually improved, demonstrating the effectiveness of our proposed components. 
It is worth noting that comparing A4 to A3, a big improvement is achieved by applying the category graphs.
This verifies that the category graphs indeed help our EPG-MGCN learn to distinguish the different motion patterns of heterogeneous traffic agents.
After adding the category graphs, our multi-graph convolutional module pays more attention to learn the similarities in the same type of traffic agents, which results in a better performance.

\begin{table}
  \caption{Ablative comparison on the ApolloScape dataset. Data are in meters.}
  \label{TB:ablation0}
  \centering
  \begin{center}
  \begin{tabular}{c|cccccc|c}
  \toprule
  Index & $G^D$ & $G^V$ & $G^P$ & $G^C$ & PGP & CS-GRU & WSADE \\[1pt]
  \midrule 
  A1 & \underline{\textbf{\checkmark}}  & x & x & x & x & x & 1.39 \\[1pt]
  A2 & \checkmark & \underline{\textbf{\checkmark}} & x & x & x & x & 1.33 \\[1pt]
  A3 & \checkmark & \checkmark & \underline{\textbf{\checkmark}} & x & x & x &  1.30\\[1pt]
  A4 & \checkmark & \checkmark & \checkmark & \underline{\textbf{\checkmark}} & x & x &  1.18\\[1pt]
  A5 & \checkmark & \checkmark & \checkmark & \checkmark & \underline{\textbf{\checkmark}} & x &  1.11\\[1pt]
  \midrule
  A6 & \checkmark & \checkmark & \checkmark & \checkmark & \checkmark & \underline{\textbf{\checkmark}} & \textbf{0.96} \\[1pt]
  \bottomrule
  \end{tabular}
  \end{center}
\end{table}

\subsection{Vehicle Trajectory Prediction}
We further evaluate our model in complex highway environments to verify its robustness and flexibility.
\subsubsection{Dataset and Experimental Settings} The NGSIM dataset contains two subdatasets for vehicle trajectories in highway: I-80 and US-101, each of which contains 45 minutes of vehicle trajectory data in different traffic conditions: mild, moderate, and congested. 
All trajectories are collected as discrete points with a frequency of 10 Hz.
For a fair comparison, we follow prior process strategy in \cite{deo2018convolutional,zhao2019multi,li2019grip}: we first downsample the raw vehicle trajectory data to 5 Hz, and then divide each trajectory into 8-second segment. We use the first 3 second as the observation time horizon, and the remaining 5 seconds as prediction time horizon.
For the ego vehicle, we consider its real driving trajectory in the prediction time horizon as the planned future motions for training.

Following existing works \cite{deo2018convolutional,li2019grip}, we assume the ego vehicles can observe the motions of vehicles within the range of $\pm$ 90 feet longitudinally and two adjacent lanes laterally.
The ego vehicle will predict the future trajectories of all observed vehicles within this region.
The same as ApolloScape, we set the distance threshold $d_D$ to be 10 meters and $\beta=20^{\circ}$ to construct the distance graphs and planning graphs. 
Since NGSIM only contains vehicle trajectories, the category graphs are not used. 
The batch size is set to be 128, and we train the model using Adam optimizer with an initial learning rate of 0.001. The learning rate is multiplied by 0.1 every 5 epochs until the convergence of the loss.

\subsubsection{Baselines} 
We evaluate the performance of EPG-MGCN by comparing with the following well-known baselines:

\begin{itemize}
  \item Constant Velocity (denoted as \textbf{CV}) \cite{deo2018convolutional}: This baseline uses a constant velocity Kalman filter to predict the future trajectory of a vehicle. This is the simplest prediction model without considering the impacts of surrounding vehicles on the target vehicle.
  \item Vanilla LSTM (denoted as \textbf{V-LSTM}) \cite{zhao2019multi}: This baseline is an encoder-decoder LSTM network that predicts a deterministic future trajectory for the target vehicle. This baseline also ignores the effects of surrounding vehicles on the target vehicle.
  \item \textbf{CS-LSTM} \cite{deo2018convolutional}: This baseline is a LSTM-based model. It proposes convolutional social pooling layers to capture the spatial interactions, and predicts the maneuver-based multi-modal trajectory distributions for the target vehicle.
  \item \textbf{MATF} \cite{zhao2019multi}: This baseline employs two parallel streams to encode scene image and historical trajectories, respectively. Then, a LSTM-based decoder is used to predict future trajectories. This baseline uses conditional generative adversarial training to capture the multimodal uncertainty.
  \item \textbf{GRIP} \cite{li2019grip}: This baseline uses the graph representation to model the interactions among vehicles and uses LSTM to predict future trajectories.
  \item Graph-based spatial-temporal convolutional network (denoted as \textbf{GSTCN}) \cite{sheng2022graph}: This baseline introduces weighted edges to the graph to capture spatial-temporal interactions and generates probability distributions over future trajectories.
  \item Planning-informed trajectory prediction model (denoted as \textbf{PiP}) \cite{song2020pip}: This baseline is the first to consider the planning information of the ego vehicle for the trajectory prediction. It uses social pooling, LSTM, and convolutional layers to learn vehicular interactions and predict maneuver-based future trajectories.
  \item Dual-attention mechanism trajectory prediction method coupled with ego motion trend (denoted as \textbf{DAMT}) \cite{guo2022vehicle}: This baseline improves PiP by introducing two attention mechanisms, i.e., spatial attention and temporal attention, to help distinguish the contributions of trajectories at different locations and times to the prediction.
\end{itemize}

\begin{table*}
  \caption{Comparison with baseline on the NGSIM dataset. The best results are marked in bold. Data are in meters.}
  \label{TB:comp}
  \centering
  \begin{center}
  \begin{tabular}{p{70pt}<{\centering}p{42pt}<{\centering}p{42pt}<{\centering}p{42pt}<{\centering}p{42pt}<{\centering}p{42pt}<{\centering}p{42pt}<{\centering}}
  \toprule
  Method & FDE@1s & FDE@2s & FDE@3s & FDE@4s & FDE@5s & ADE \\[1pt]
  \midrule 
  CV & 0.73 & 1.78 & 3.13 & 4.78 & 6.68 & 3.42 \\[1pt]
  V-LSTM & 0.66 & 1.62 & 2.94 & 4.63 & 6.63 & 3.30 \\[1pt]
  CS-LSTM & 0.58 & 1.32 & 2.22 & 3.26 & 4.40 & 2.36 \\[1pt]
  MATF & 0.66 & 1.34 & 2.08 & 2.97 & 4.13 & 2.24 \\[1pt]
  GRIP & 0.64 & 1.13 & 1.80 & 2.62 & 3.60 & 1.96 \\[1pt]
  GSTCN & 0.44 & 0.83 & 1.33 & 2.01 & 2.98 & 1.52 \\[1pt]
  PiP & 0.55 & 1.19 & 1.95 & 2.90 & 4.07 & 2.13 \\[1pt]
  DAMT & 0.50 & 1.11 & 1.78 & 2.69 & 3.93 & 2.00 \\[1pt]
  \midrule
  EPG-MGCN (ours) & \textbf{0.29} & \textbf{0.76} & \textbf{1.32} & \textbf{1.97} & \textbf{2.72} & \textbf{1.41} \\[1pt]
  \bottomrule
  \end{tabular}
  \end{center}
\end{table*}

\subsubsection{Comparison of Prediction Results}
Table \ref{TB:comp} shows the prediction results of each model in terms of ADE and FDEs at different prediction time steps. We can see that our proposed EPG-MGCN achieves the best performance in terms of the ADE and all FDEs. 
According to the results in Table \ref{TB:comp}, we draw the following conclusions.

Similar to our proposed network, PiP and DAMT also consider the future ego-vehicle planning information. Compared with them, our network have better prediction results on ADE and all FDEs, and achieves about 34\% and 30\% performance improvement in terms of ADE, respectively. This demonstrates that the graph convolution is more effective to capture the social interactions than LSTM.
When compared with GRIP and GSTCN, our EPG-MGCN also outperforms them. This verifies that the proposed planning-guided prediction and multi-graph describing the influences from multiple perspectives are indeed helpful for modeling the complex social interactions between vehicles.

It is worth noting that even without taking the scene images as additional inputs, our method still outperforms those methods utilizing the scene context features, such as MATF. 
This illustrates the superiority of graph convolution over LSTM and CNN in learning interactive relations, and informs us that the performance of our proposed EPG-MGCN could potentially be further improved by considering the scene context images.

\begin{figure*}
  \centerline{\includegraphics[width=17cm]{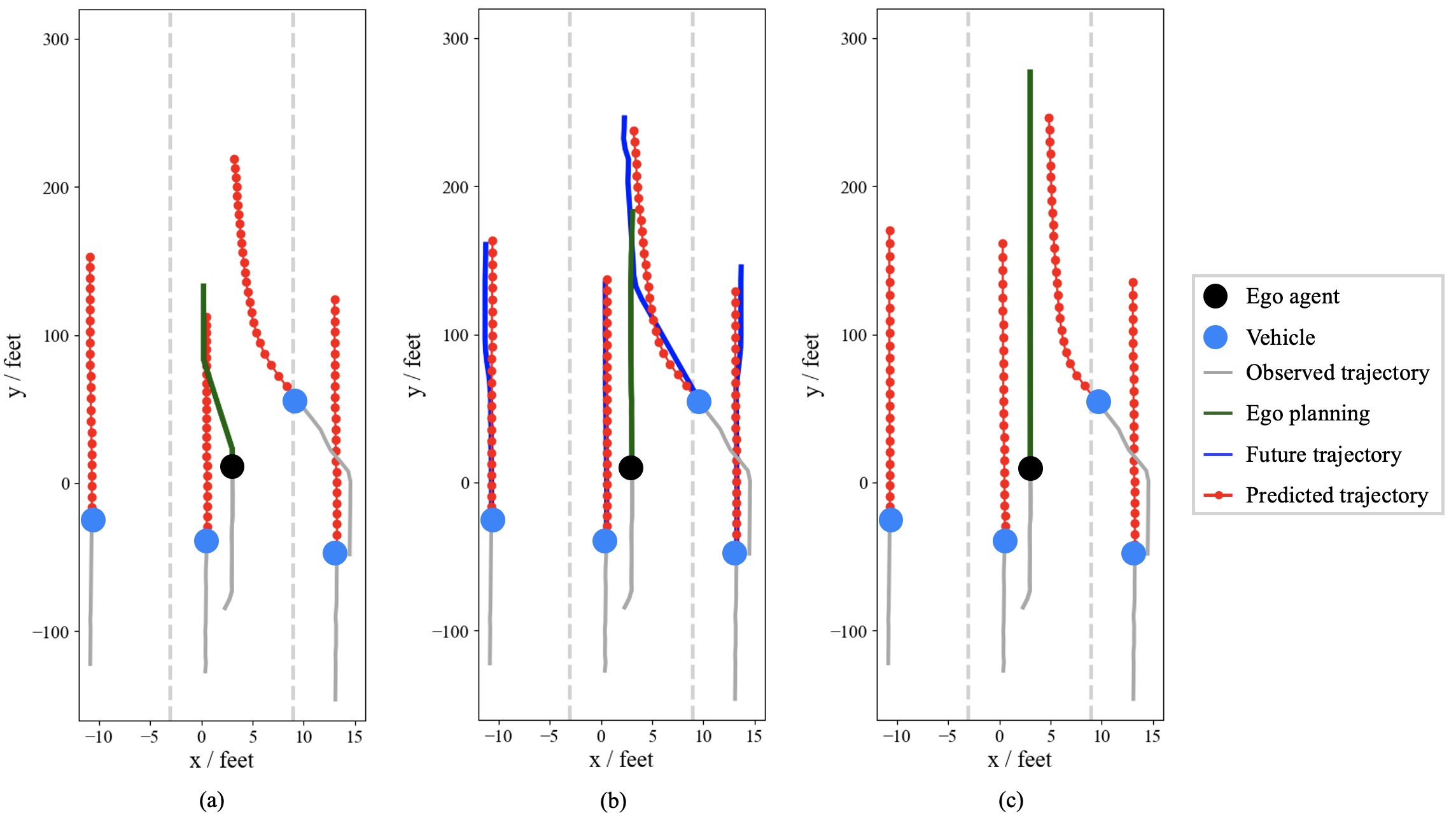}}
  \caption{Prediction under different ego-agent plans: (a) the ego vehicle decelerates and turns to the left; (b) the ego vehicle keeps its current driving speed and direction; (c) the ego vehicle speeds up.}
  \label{FG:visual_plans}
\end{figure*}

\subsection{Prediction under Different Ego-Agent Plans} 
To investigate if our proposed EPG-MGCN has learned to distinguish the impacts of ego-agent plans on the future motions of nearby traffic agents, we predict the future scenarios under different planned trajectories executed by the ego vehicle.
Fig. \ref{FG:visual_plans} shows predicted trajectories, where the observed trajectories are acquired from the NGSIM dataset. 
After obtaining 3 seconds of observable trajectories and the future ego-agent plans, our network predicts their trajectories over 5 seconds in the future. 
We find that given different ego-agent plans, our EPG-MGCN predicts a diverse set of future scenes.

As shown in Fig. \ref{FG:visual_plans}(a), the proposed EPG-MGCN predicts that the left-rear vehicle would decelerate to avoid a collision if the ego vehicle plans to decelerate and turn to the left.
The ego vehicle in Fig. \ref{FG:visual_plans}(b) executes the future trajectory acquired from the NGSIM dataset, so we compare the predicted trajectories with the ground truth future trajectories.
We can see that our model accurately predicts the future trajectories for nearby cars. In addition, our model successfully predicts that the right-preceding car is going to change to the middle lane.
In Fig. \ref{FG:visual_plans}(c), the ego vehicle decides to accelerate to overtake the right-preceding car. 
Our model predicts that the right-preceding car would execute a smaller lateral displacement than that in Fig. \ref{FG:visual_plans}(b) to ensure safety.
From the results in Fig. \ref{FG:visual_plans}, we find that our proposed EPG-MGCN can generate diverse and reasonable predictions for future situations, which is highly beneficial for autonomous driving in interactive scenarios. 

\section{Conclusions and Future Work}
In this paper, we have presented an ego-planning guided multi-graph convolutional network (EPG-MGCN) to predict the future trajectories of heterogeneous target agents around the ego vehicle. In our method, a multi-graph convolutional module has been proposed to distinguish different impacts of agents on each other from four perspectives, i.e., distance, visibility, ego-planning, and category. Based on these four graph topologies, EPG-MGCN can well model and learn the social interactions among heterogeneous traffic agents. Further, we have proposed a planning-guided prediction module to leverage the fact that the future motions of surrounding agents are impacted by the future planning of the ego vehicle. The experimental results have shown that the proposed EPG-MGCN outperforms the mainstream methods, demonstrating its potential to be deployed in autonomous vehicles.

The proposed network can be further improved in several directions. Firstly, in addition to trajectories, many additional data (e.g., visual images, sound signals, high-resolution maps, vehicle-to-vehicle/infrastructure communication information) can be collected and used by autonomous vehicles. Therefore, how to incorporate these data to enhance prediction performance is worth studying. 
Secondly, when facing a traffic scene, people may have multiple choices. Therefore, another promising direction is to introduce multi-modal prediction to handle the randomness of traffic agents' behaviors. 

\ifCLASSOPTIONcaptionsoff
  \newpage
\fi

{\small
\bibliographystyle{IEEEtran}
\bibliography{IEEEabrv,egbib}
}

\end{document}